\documentclass[10pt,twocolumn,letterpaper]{article}

\usepackage[utf8]{inputenc}
\usepackage{microtype}
\usepackage[margin=0.75in,top=0.9in,bottom=0.9in,columnsep=0.25in]{geometry}
\usepackage{amsmath,amssymb}
\usepackage{graphicx}
\usepackage{booktabs}
\usepackage{array}
\usepackage{multirow}
\usepackage{caption}
\usepackage{algorithm}
\usepackage{enumitem}
\usepackage{cite}
\usepackage{url}
\usepackage[colorlinks=true,linkcolor=blue,citecolor=blue,urlcolor=blue,
            pdftitle={Parameter misspecification in physics-informed neural networks: A closed-form audit and its limits},
            pdfauthor={David McShannon, Nicholas Dietrich}]{hyperref}

\newcommand{\norm}[1]{\left\|#1\right\|}
\newcommand{\est}[1]{\widehat{#1}}
\newcommand{\se}{\mathrm{se}_{\mathrm{boot}}}
\newcommand{\stated}{\mathrm{stated}}
\newcommand{\trained}{\mathrm{trained}}
\renewcommand{\Re}{\mathit{Re}}
\newcommand{\Pe}{\mathit{Pe}}
\newcommand{\dlog}{d\log}

\title{Parameter misspecification in physics-informed neural networks:\\[2pt] A closed-form audit and its limits}

\author{David McShannon \and Nicholas Dietrich}

\date{}

\begin{document}

\twocolumn[
\begin{@twocolumnfalse}
\maketitle
\begin{abstract}
\noindent
Physics-informed neural networks (PINNs) are trained to minimize the residual of a prescribed equation, so a low training loss does not indicate whether a coefficient in that equation was specified incorrectly. This study evaluates what can be inferred about the encoded parameter from the trained network alone. When the residual is linear in the parameter, written as $A - \nu B$, the least-squares estimate $\est{\nu} = A^{\top}B/B^{\top}B$ is obtained from a single evaluation of the frozen network, without retraining or reference data. Its accuracy depends on how closely the network satisfies its own equation, quantified by a normalized residual $\rho$. Recovery error is associated with $\rho$ (log-log correlation 0.72 across 36 convection-diffusion networks), and empirical exponents of 0.63 to 1.20 across three equations decompose into one half plus a measurable angular term. For Burgers' equation, a single-model test based on a block-bootstrap standard error flagged 3 of 300 correctly specified networks (observed false-alarm rate 1\%), all 2\% misspecifications at a practical budget, and two thirds of 1\% misspecifications after convergence. For the self-similar blowup rate of the C\'ordoba--C\'ordoba--Fontelos equation, where the rate also enters the architecture, the estimator recovers about half of the misspecification, whereas $\rho$ increases approximately quadratically and distinguishes a 0.2\% error. When the incorrect rate enters both the equation and the architecture, errors of up to 2\% were not detected at the precision reached. At practical budgets, the raw loss can be lower for the misspecified model, a scaling effect that $\rho$ does not share.
\end{abstract}
\vspace{1.5em}
\end{@twocolumnfalse}
]

\section{Introduction}

A physics-informed neural network (PINN) is trained by minimizing the residual of a governing equation, so the prescribed equation defines the optimization objective \cite{r1}. PINNs have been built into digital-twin prognostics for hydraulic systems \cite{r2}, used to reconstruct the flow in transonic turbine cascades \cite{r3} and personalized to cardiac data \cite{r4}. In such settings the coefficients of the equation come from a configuration file, an upstream simulation or a collaborator, and a low training loss may be taken as confirmation that they are correct.

This interpretation presupposes a correctly specified equation, because the loss is computed against the equation used in training. If a coefficient is incorrect because of misconfiguration, a unit mismatch or an outdated value, the network can solve the incorrect problem accurately while the loss indicates successful training, and at practical budgets the raw loss can even be lower for the misspecified model. This study evaluates what can be determined from the trained model alone: which parameter the model encodes, how accurately that parameter can be recovered from the frozen weights, and under what conditions the recovery fails. The last question is examined in the context of PINN-based singularity discovery \cite{r5}, where the residual is one part of a validation pipeline whose other parts exist because the residual alone is not sufficient.

The first question admits a closed-form answer, because a misspecified network approximately solves the equation used in training, and the coefficient of that equation can be recovered from the network. Writing the residual as $A - \nu B$, with $A$ the terms that do not involve the coefficient $\nu$ and $B$ the terms it multiplies, the value of $\nu$ that minimizes the residual of a frozen network is
\begin{equation}
\est{\nu} = \frac{A^{\top}B}{B^{\top}B}.
\label{eq:1}
\end{equation}
It requires one evaluation of the network and two inner products, and comparing it with the documented parameter gives a consistency check that needs no reference model, no retraining and no solution data. The estimate is exact only when the network satisfies its equation exactly, which a trained PINN does not achieve in practice, and its accuracy depends on a normalized residual that can itself be computed from the trained model.

The contributions of this paper are as follows.
\begin{enumerate}[leftmargin=*,itemsep=2pt,topsep=3pt]
\item We propose Eq.~(\ref{eq:1}), the single-column case of equation-discovery regression \cite{r6,r7}, as an audit of the coefficient documented for a trained PINN, and pair it with a spatial block-bootstrap standard error and a decision rule that require only the trained model.
\item We express the relative recovery error exactly in terms of the normalized residual and the angle between residual and basis, and we measure, for Burgers' equation, a lid-driven cavity and convection-diffusion, empirical exponents that decompose into one half plus an angular drift that differs by system and by parameter.
\item We calibrate the decision rule on Burgers' equation against 300 correctly specified networks and report its false-alarm rate and the smallest misspecification it detects at a practical and at a converged training budget.
\item We demonstrate that misspecified networks can reach training losses at or below those of correctly specified ones, and that at practical budgets the raw loss can favor the misspecified model while the normalized residual does not.
\item We examine the self-similar blowup rate of the C\'ordoba--C\'ordoba--Fontelos equation, where the rate enters the network architecture as well as the equation, and identify what the estimator and the residual can and cannot establish in that setting.
\end{enumerate}

\section{Related work}

Recovering PDE coefficients by least squares on derivatives of a learned field is established. PDE-FIND \cite{r6} regresses a library of derivative terms against the time derivative, and DL-PDE \cite{r7} obtains the derivatives by automatic differentiation of a trained network, which is what Eq.~(\ref{eq:1}) does with the library reduced to one known column. Inverse PINNs \cite{r1} treat the parameter as trainable and fit it jointly with the solution against data. Eq.~(\ref{eq:1}) instead interrogates a network whose weights are frozen and audits a documented coefficient, so it is assessed by the smallest discrepancy it resolves at a stated false-alarm rate.

A posteriori methods bound the solution error of a neural network approximation from computable residual quantities \cite{r8} or certify the worst-case residual of a PINN over its domain \cite{r9}. Gorard et al.~\cite{r10} construct neural solvers for hyperbolic equations with extrapolatory $L^{\infty}$ error bounds, and Baheri and Laguna Peralta \cite{r11} compile small Fourier neural operators into SMT to check positivity and mass conservation. These methods certify properties of the computed solution. Shikhman \cite{r12} audits learned simulators with structural diagnostics that require reference trajectories and reports that state error can improve while those diagnostics deteriorate. We are not aware of an earlier residual-to-parameter result for PINNs, and its form is the backward-error analysis of least squares \cite{r13,r14,r15,r16} applied to the PINN residual.

Wang et al.~\cite{r5} use PINNs to find unstable self-similar singularities, reaching residuals near machine precision with Gauss--Newton and multistage training \cite{r17}, and validate the blowup rate by requiring it to minimize the attainable residual, which involves retraining at each candidate rate. Other known difficulties of PINN training include spectral bias \cite{r18}, ill-conditioned loss landscapes \cite{r19} and overoptimistic reporting \cite{r20}. Misspecification of the equation itself has not been characterized for PINNs.

\section{Method}

\subsection{Estimator}

A PINN approximates the solution of $\mathcal{N}[u;\lambda] = 0$ in $\Omega$ with boundary and initial conditions by minimizing the sum of the mean squared interior, boundary and initial residuals, here with equal weights. The network is misspecified when trained with $\lambda' = \lambda(1 + \delta)$ while $\lambda$ is the documented value, and the analyst is assumed to have access only to the trained model, its loss and the documented parameter. When the interior residual is linear in the parameter it can be written as $A - \nu B$ on a set of interior points, and $\norm{A - \nu B}^{2}$ is minimized by Eq.~(\ref{eq:1}). Table~\ref{tab:1} gives $A$ and $B$ for the systems studied. Because the squared residual norm is quadratic in $\nu$, Eq.~(\ref{eq:1}) is the continuous minimizer of a parameter sweep. With $d$ parameters it becomes $\est{\mathbf{p}} = \left( M^{\top}M \right)^{-1}M^{\top}A$ with $M = \left[ B_{1}\ \cdots\ B_{d} \right]$. Two quantities are reported, the recovery offset $\left| \est{\nu} - \nu_{\stated} \right|/\nu_{\stated}$ against the documented value, which can be computed in practice, and the recovery error $\left| \est{\nu} - \nu_{\trained} \right|/\nu_{\trained}$ against the value used in training, which measures the accuracy of the estimator.

\begin{table*}[t]
\caption{Decomposition of the interior residual as $A - \nu B$ for each system. For the cavity the two momentum residuals are stacked into one vector and the pressure gradient is included in $A$.}
\label{tab:1}
\centering
\small
\begin{tabular}{llll}
\toprule
System & Residual & $A$ & $B$ \\
\midrule
Burgers & $u_{t} + uu_{x} - \nu u_{xx}$ & $u_{t} + uu_{x}$ & $u_{xx}$ \\
Cavity ($\nu = 1/\Re$) & $\left( \mathbf{u} \cdot \nabla \right)\mathbf{u} + \nabla p - \nu\nabla^{2}\mathbf{u}$ & $\left( \mathbf{u} \cdot \nabla \right)\mathbf{u} + \nabla p$ & $\nabla^{2}\mathbf{u}$ \\
Convection-diffusion & $u_{t} + \mathbf{c} \cdot \nabla u - D\nabla^{2}u$ & $u_{t} + \mathbf{c} \cdot \nabla u$ & $\nabla^{2}u$ \\
CCF profile (rate $\lambda$) & $\lambda\Theta - (1 + \lambda)y\Theta' + s\left( \mathcal{H}\Theta \right)\Theta'$ & $- y\Theta' + s\left( \mathcal{H}\Theta \right)\Theta'$ & $y\Theta' - \Theta$ \\
\bottomrule
\end{tabular}
\end{table*}

\subsection{Normalized residual and error decomposition}

How well a frozen network satisfies its own equation is measured by
\begin{equation}
\rho = \frac{\norm{A - \est{\nu}B}^{2}}{\norm{A}^{2}},
\label{eq:2}
\end{equation}
which, unlike the raw residual, is comparable across training stages and systems. Since $\est{\nu}$ is the least-squares value, $\rho$ is the squared sine of the angle between $A$ and $B$. Let $r = A - \nu B$ be the residual at the true parameter. Then $\est{\nu} - \nu = r^{\top}B/B^{\top}B$, and $A - \est{\nu}B$ is the component of $r$ orthogonal to $B$. With $\theta$ the angle between $r$ and $B$,
\begin{equation}
\varepsilon \equiv \frac{\left| \est{\nu} - \nu \right|}{\nu} = \frac{\norm{A} \sqrt{\rho}\,|\cot\theta|}{\nu\, \norm{B}}.
\label{eq:3}
\end{equation}
This identity fixes the analytic exponent between $\varepsilon$ and $\rho$ at one half. Writing $K = \norm{A}/(\nu \norm{B})$, the exact logarithmic slope is
\begin{equation}
\frac{\dlog\varepsilon}{\dlog\rho} = \frac{1}{2} + \frac{\dlog|\cot\theta|}{\dlog\rho} + \frac{\dlog K}{\dlog\rho}.
\label{eq:4}
\end{equation}
When the network nearly satisfies its equation, $K \approx 1$, and in every system measured here $\log K$ varies by less than 0.002 across cells, so the last term is small and the exponent observed in a log-log fit is approximately one half plus the angular term. The size of the omitted term is reported alongside the decomposition in Section~\ref{sec:5.3}.
In least-squares backward-error analysis \cite{r13,r15,r16} the gap between relative residual and parameter error is likewise set by the angle between residual and column space and can be arbitrarily loose \cite{r14}. The present study measures where PINN residuals lie relative to the tight case.

\subsection{Uncertainty and decision rules}

The homoskedastic standard error of Eq.~(\ref{eq:1}) is optimistic \cite{r21} because PINN residuals concentrate near shocks and corners, so a spatial block bootstrap \cite{r22,r23,r24} is used instead. The evaluation points are divided into 64 tiles by coordinate, tiles are resampled with replacement, $\est{\nu}$ is recomputed from the stored $A$ and $B$, and the standard deviation over 400 resamples defines $\se$. The single-model rule flags a network when $z = \left| \est{\nu} - \nu_{\stated} \right|/\se > 2.576$, the two-sided normal threshold at 1\%. The population rule flags a network whose offset exceeds the empirical 99th percentile of offsets from correctly specified networks and is used to check the calibration of the single-model rule. Tiles must be formed on coordinates and not on index order, because consecutive points of a low-discrepancy sequence are spread over the whole domain and index-based blocks would reduce the procedure to an independent bootstrap. Algorithm~\ref{alg:1} summarizes the procedure. It costs one forward pass with the required derivatives, the work of one loss evaluation, and each bootstrap resample costs $O(N)$ without further network evaluation, whereas a $k$-point sweep needs $k$ residual evaluations and a retrain-per-value search $k$ trainings.

\begin{algorithm}[t]
\caption{Closed-form audit of the encoded parameter.}
\label{alg:1}
\small
\emph{Input:} trained network $\est{u}$, stated parameter $\nu_{\stated}$, interior points $\{ x_{i}\}_{i = 1}^{N}$ and a partition of them into 64 coordinate tiles.
\begin{enumerate}[leftmargin=*,itemsep=1pt,topsep=2pt]
\item Evaluate $\est{u}$ and its derivatives at $\{ x_{i}\}$ by automatic differentiation and form $A,B \in \mathbb{R}^{N}$.
\item Compute $\est{\nu} = \left( A^{\top}B \right)/\left( B^{\top}B \right)$ and $\rho = \norm{A - \est{\nu}B}^{2}/ \norm{A}^{2}$.
\item For $m = 1,\ldots,M_{\mathrm{boot}}$, draw 64 tiles with replacement, let $S$ be the union of their indices, and compute $\est{\nu}_{m} = \left( A_{S}^{\top}B_{S} \right)/\left( B_{S}^{\top}B_{S} \right)$.
\item Set $\se = \mathrm{sd}\left( \est{\nu}_{1},\ldots,\est{\nu}_{M_{\mathrm{boot}}} \right)$ and $z = \left| \est{\nu} - \nu_{\stated} \right|/\se$.
\item Return $\est{\nu}$, $\rho$, $\se$ and the flag $z > 2.576$.
\end{enumerate}
\end{algorithm}

\section{Test problems and experimental design}

\subsection{Test problems}

Burgers' equation is solved at $\nu = 0.01/\pi$ in the configuration of Raissi et al.~\cite{r1}, with the exact solution from the Cole--Hopf transformation \cite{r25,r26}. The lid-driven cavity is solved at $\Re = 100$ and 400 with reference data from Ghia et al.~\cite{r27}, using either the standard lid, whose velocity is discontinuous at the upper corners \cite{r28}, or the smoothed lid $u(x,1) = 1 - \left( (2x - 1)^{18} \right)^{2}$, which vanishes at both corners and exceeds 0.99 over 88\% of the lid. Convection-diffusion is solved as a two-dimensional Gaussian pulse at P\'eclet number 33.5, for which the analytic solution of the pulse serves as reference, and as a one-dimensional pulse in which $D$ is varied to give P\'eclet numbers from 10 to 200.

The fourth problem is the self-similar profile of the C\'ordoba--C\'ordoba--Fontelos (CCF) model \cite{r29}, which Wang et al.~\cite{r5} use to test PINN-based singularity discovery. With $\phi(x,t) = \phi_{0} - (1 - t)^{\lambda}\Theta(y)$ and $y = x/(1 - t)^{1 + \lambda}$, the profile equation is
\begin{equation}
\lambda\Theta - (1 + \lambda)\, y\,\Theta' + s\,\left( \mathcal{H}\Theta \right)\,\Theta' = 0,
\label{eq:5}
\end{equation}
with $\mathcal{H}$ the Hilbert transform and $s = \pm 1$. Smooth solutions exist only at isolated rates, the stable one being $\lambda_{0} = 1.1807776628998$ \cite{r5}. Following Wang et al., the network represents $\omega = \Theta'$ in a compactified coordinate with a decay envelope, and both the coordinate and the envelope depend on $\lambda$, so the rate enters the architecture as well as the equation.

\subsection{Networks and training}

All networks are fully connected tanh networks, except one sine-activated variant in the architecture comparison, implemented in PyTorch \cite{r30}, using DeepXDE \cite{r31} for the three benchmark problems, trained with Adam \cite{r32} followed where stated by L-BFGS \cite{r33,r34}, with Glorot initialization \cite{r35}. The benchmark problems use single precision and the detection-floor and CCF experiments double precision. Weights are stored at checkpoints along each run. Adam followed by L-BFGS is a common training strategy for PINNs \cite{r36}. The checkpoint at the end of the Adam phase is called the practical budget, and the checkpoint after L-BFGS the converged budget. Supplementary Table~\ref{tab:S1} lists architectures, collocation, optimizer settings and seed counts for every experiment. $A$ and $B$ are evaluated on 10,000 scrambled Sobol' points \cite{r37,r38}, excluding a strip of width $h_{w} = 0.02$ along the cavity walls.

\subsection{Experiments}

Misspecified networks are trained at $\delta$ from 5 to 200\% at the practical budget and compared with correctly specified networks through the training loss and the relative $L^{2}$ distance between predictions. Solution differences are judged against a drift floor, the distance between correctly specified networks that differ only in seed, computed over all pairs of seven clean and seven misspecified Burgers networks at $\delta = 50\%$ and over the corresponding cavity networks. For the cavity the loss is additionally recorded for perturbations below and above $\Re = 100$, for five architectures at $\delta = 50\%$, and per loss term at 25k, 50k and 100k Adam steps and after L-BFGS. Six loss-based checks are evaluated, namely the residual against the stated equation, an ensemble of misspecified models, an ensemble trained at $\pm 2$, $\pm 5$ and $\pm 10\%$ around the stated value, a gradient-based fit of the parameter on the frozen network, the divergence of the velocity, and loss elevation against a clean baseline with threshold 0.01.

Recovery is evaluated with Eq.~(\ref{eq:1}) on all problems, and for the cavity it is compared with a 31-point sweep over $\Re$ for networks trained at $\Re = 100$ and at $\Re = 150$ while documented as 100. The detection floor is measured on Burgers' equation with 300 correctly specified networks at the practical budget, 30 of which are continued through L-BFGS, and six misspecified networks per level at $\delta$ between $\pm 0.5$ and $\pm 5\%$. Normality of the clean offsets is tested with the Shapiro--Wilk test \cite{r39}, and stationarity of the converged estimates by logging $\est{\nu}$ every 1,000 L-BFGS iterations. Cavity models trained at $\Re = 400$, and at $\Re = 600$ while documented as 400, are also compared with the 17-point centerline data of Ghia et al.~\cite{r27}, and the wall-strip width $h_{w}$ is varied from 0 to 0.10.

The relation between recovery accuracy and $\rho$ is measured on the cavity with both lids and with L-BFGS budgets from 50k to 300k iterations, on Burgers' equation, and on a P\'eclet sweep from 10 to 100 with three perturbation levels, three seeds and three checkpoints. Exponents are slopes of ordinary least-squares fits of log recovery error on $\log\rho$, with uncertainty from a bootstrap over networks, and $\cot\theta$ is reconstructed from stored $A$ and $B$ to evaluate Eq.~(\ref{eq:4}). Joint recovery of $c$ and $D$ is tested in fifteen runs and over P\'eclet numbers from 10 to 200, and the bootstrap standard error is compared with the scatter across seeds. For the CCF profile, a model documented as using $\lambda_{0}$ has an architecture built from $\lambda_{0}$, so in the main experiment only the residual is evaluated at $\lambda_{0}(1 + \delta)$, with $\delta$ from $- 2$ to $5\%$, three seeds per level and thirty clean networks for calibration. In a second set of 24 networks the architecture is also built from $\lambda_{0}(1 + \delta)$, with $\delta$ from 0.1 to 5\%. The clean CCF profiles are checked independently through the asymptotic constant of $\omega$ and through the relation for $\lambda$ obtained by requiring the second derivative of the residual to vanish at the origin.

\section{Results}

\subsection{Training loss under misspecification}
\label{sec:5.1}

At the practical budget, misspecified models for Burgers' equation and for the cavity at $\Re = 400$ reach training losses at or below the correctly specified baseline while their solutions differ substantially from it (Supplementary Table~\ref{tab:S2}). For the cavity at $\Re = 100$ and for convection-diffusion the misspecified losses are higher than the baseline, by factors of up to about four. For the cavity at $\Re = 400$, every perturbation from 5 to 100\% gives a solution difference of 26 to 71\% at a loss within seed scatter of the baseline (0.0071 to 0.0089 against 0.0088). For Burgers the PINN solutions differ by 0.6 to 8.0\% where the exact solutions differ by 0.25 to 7.3\%, and for convection-diffusion the PINN differences of 1.9 to 40\% compare with 1.5 to 35\% for the analytic pulse. The misspecified networks therefore approximate, within seed scatter, the solution of the equation they were given, and the size of the solution difference reflects the sensitivity of each equation to its coefficient.

Raw loss and the normalized residual order the models differently during training (Table~\ref{tab:2}). With $\delta = 50\%$ at $\Re = 400$ the misspecified cavity has the lower raw loss after 25k and 100k Adam steps, separating completely over three seeds ($0.00626 \pm 0.00073$ against $0.00750 \pm 0.00038$ at 25k), while $\rho$ is larger for the misspecified model at every checkpoint (at 25k, $3.46 \pm 0.54$ against $3.24 \pm 0.50 \times 10^{- 2}$, within seed scatter). The per-term losses show that every raw component, including the boundary terms that make up 80 to 84\% of the total, is lower for the misspecified model. Only the viscous part of the momentum residual carries $\nu$, so a $\nu$ smaller by a factor of 2/3 bounds the momentum ratio between 4/9 and 1, and the observed 0.66 lies within these bounds. Raw loss therefore cannot be compared across parameter values. After L-BFGS the ordering reverses in every component and $\rho$ is four times larger for the misspecified model, which is attributed to the $\Re = 600$ field converging less well under the standard lid (Section~\ref{sec:5.3}).

\begin{table*}[t]
\caption{Cavity at $\Re = 400$ through training, three networks per arm, $\delta = 50\%$, standard lid, $[50] \times 4$ network. Raw training loss inverts at two checkpoints and the normalized residual at none. Values are means over three seeds; standard deviations at the 25k checkpoint are given in Section~\ref{sec:5.1}.}
\label{tab:2}
\centering
\small
\begin{tabular}{lcccccc}
\toprule
Checkpoint & Raw loss, Clean & Raw loss, Misspecified & Raw loss, Inverted & $\rho$, Clean & $\rho$, Misspecified & $\rho$, Inverted \\
\midrule
25k Adam & 0.00750 & 0.00626 & yes & 0.0324 & 0.0346 & no \\
50k Adam & 0.00589 & 0.00515 & partly & 0.0260 & 0.0263 & no \\
100k Adam & 0.00658 & 0.00399 & yes & 0.0246 & 0.0272 & no \\
L-BFGS & 0.00025 & 0.00058 & no & 0.0026 & 0.0112 & no \\
\bottomrule
\end{tabular}
\end{table*}

Solution differences are also difficult to attribute at practical budgets. For Burgers at $\delta = 50\%$ the distance between misspecified and clean networks is 0.79 times the drift floor after 20k Adam steps and 1.49 times after L-BFGS, and even then the two distributions overlap (Supplementary Table~\ref{tab:S3}). Six checks based on the loss, including ensembles, a gradient-based fit of the parameter and the divergence of the velocity, are each unreliable in at least one regime (Supplementary Table~\ref{tab:S4}). In most cases, misspecification was not reflected in the loss for perturbations in either direction (Supplementary Table~\ref{tab:S5}), and whether it raises the loss depends on the architecture (Supplementary Table~\ref{tab:S6}).

\subsection{Recovery and detection floor}
\label{sec:5.2}

\begin{figure}[t]
\centering
\includegraphics[width=\columnwidth]{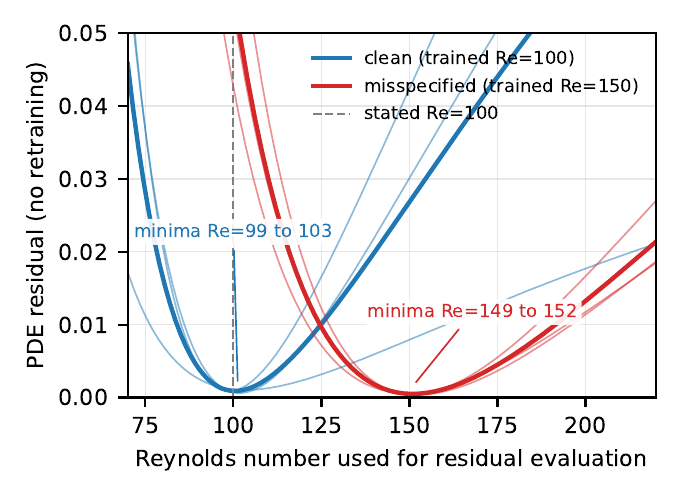}
\caption{Parameter sweep for the cavity. The PDE residual of each frozen network is evaluated over a range of $\Re$ without retraining, for three seeds. Thin lines show individual seeds and thick lines the mean. The mean minimum of the correctly specified networks falls at the stated value $\Re = 100$, and that of the misspecified networks at their training value $\Re = 150$. For a frozen network the squared residual norm is exactly quadratic in $\nu = 1/\Re$, so the curves are the parabolas through the 31-point sweeps. Closed-form minima are $\Re = 99$ to 103 (clean) and 149 to 152 (misspecified).}
\label{fig:1}
\end{figure}

For the cavity the closed-form minima fall at $\Re = 99$ to 103 for correctly specified networks and at 149 to 152 for networks trained at $\Re = 150$ (Fig.~\ref{fig:1}), and for Burgers and convection-diffusion they fall at the training value to within the sweep grid step.

The detection floor was measured on double-precision Burgers networks at two checkpoints of the same runs, after 20k Adam steps and after a further 30k L-BFGS iterations, with six seeds per perturbation level (Table~\ref{tab:3}). At the practical budget 300 clean networks give offsets with median 0.167\%, 99th percentile 1.14\% and maximum 3.93\%. The distribution is heavy-tailed (normal and lognormal fits rejected at $p < 10^{- 10}$ by Shapiro--Wilk tests \cite{r39}), so the percentile is taken empirically. The single-model rule flags 3 of the 300 clean networks, an observed false-alarm rate of 1.0\% in this sample, and flags every network with $|\delta| \geq 2\%$ and 9 of 12 at $\pm 1\%$, so it reaches the 2\% floor of the population rule without a clean population. Its calibration is only approximate at this budget, since the clean signed offsets have a standard deviation of 0.39\% against a median $\se$ of 0.19\%, and a threshold of 2 instead of 2.576 would flag 5\% of clean networks. After L-BFGS the 30 clean offsets have median 0.114\% and maximum 0.742\%. Thirty networks cannot fix an empirical 99th percentile, and lognormal and normal extrapolations of it differ by a factor of two. The rule is then calibrated, with the clean standard deviation of 0.202\% matching the median $\se$ of 0.199\%, and it flags none of the 30 clean networks, which does not establish a zero false-alarm probability, two thirds of the networks at $\pm 1\%$ and all but one at $|\delta| \geq 1.5\%$.

Negative perturbations are recovered as accurately as positive ones, with mean signed offsets of $- 0.98$, $- 2.09$ and $- 2.98\%$ at $\delta = - 1$, $- 2$ and $- 3\%$ against $+ 0.99$, $+ 2.04$ and $+ 2.87\%$ on the positive side. The clean offsets carry a small bias of $- 0.12\%$, and $\est{\nu}$ is constant to three decimals in percent from about 12k L-BFGS iterations onward, so the converged values are stationary. The largest clean offsets belong to the networks with the largest $\rho$, and the logarithm of the offset correlates with $\log\rho$ at $+ 0.45$ over the 300 practical-budget networks. This is consistent with the tail of the clean distribution being determined by the quality of the optimum reached rather than by the iteration count. Against the centerline data of Ghia et al.~\cite{r27} the misspecified cavity models are worse than clean ones by factors of only 1.0 to 1.8, because the clean models already carry errors of 10 to 63\% (Supplementary Table~\ref{tab:S7}) and the standard-lid models about three times the error of the smoothed-lid models, whereas the recovery offset separates the same models completely. Recovery is insensitive to the width of the excluded wall strip between 0.01 and 0.05.

\begin{table*}[t]
\caption{Detection floor on Burgers at two budgets, double precision, six misspecified networks per level. The column $z$ counts networks with $\left| \est{\nu} - \nu_{\stated} \right|/\se > 2.576$ (single-model rule; observed clean false alarms 3/300 at the practical budget and 0/30 at the converged budget). The column $p_{99}$ counts networks above the empirical 99th percentile of the 300 clean practical-budget offsets (1.14\%). The column ``max'' counts networks above the maximum clean converged offset (0.742\% over 30 networks). ``Min.\ offset'' is the smallest offset in the group. Levels not run at a budget are marked with a dash.}
\label{tab:3}
\centering
\small
\begin{tabular}{lcccccc}
\toprule
$\delta$ & Practical, Min.\ offset & Practical, $z$ & Practical, $p_{99}$ & Converged, Min.\ offset & Converged, $z$ & Converged, max \\
\midrule
$- 3\%$ & 2.334\% & 6/6 & 6/6 & 2.373\% & 6/6 & 6/6 \\
$- 2\%$ & 1.649\% & 6/6 & 6/6 & 1.592\% & 5/6 & 6/6 \\
$- 1\%$ & 0.239\% & 4/6 & 1/6 & 0.506\% & 4/6 & 5/6 \\
$+ 0.5\%$ & -- & -- & -- & 0.150\% & 1/6 & 1/6 \\
$+ 1\%$ & 0.731\% & 5/6 & 1/6 & 0.841\% & 4/6 & 6/6 \\
$+ 1.5\%$ & -- & -- & -- & 1.174\% & 6/6 & 6/6 \\
$+ 2\%$ & 1.747\% & 6/6 & 6/6 & 1.821\% & 6/6 & 6/6 \\
$+ 3\%$ & 2.654\% & 6/6 & 6/6 & -- & -- & -- \\
$+ 5\%$ & 4.274\% & 6/6 & 6/6 & -- & -- & -- \\
\bottomrule
\end{tabular}
\end{table*}

\subsection{Residual quality and recovery accuracy}
\label{sec:5.3}

With the standard lid the cavity does not reach a low residual at any budget, because no smooth network can represent the corner discontinuity. The smoothed lid lowers the attainable $\rho$ from $5.14 \times 10^{- 3}$ to $5.72 \times 10^{- 4}$ and reduces the recovery error of the cavity from 11.9\% to 0.085\%, while increasing the L-BFGS budget from 50k to 300k iterations leaves $\rho$ unchanged.

\begin{figure}[t]
\centering
\includegraphics[width=\columnwidth]{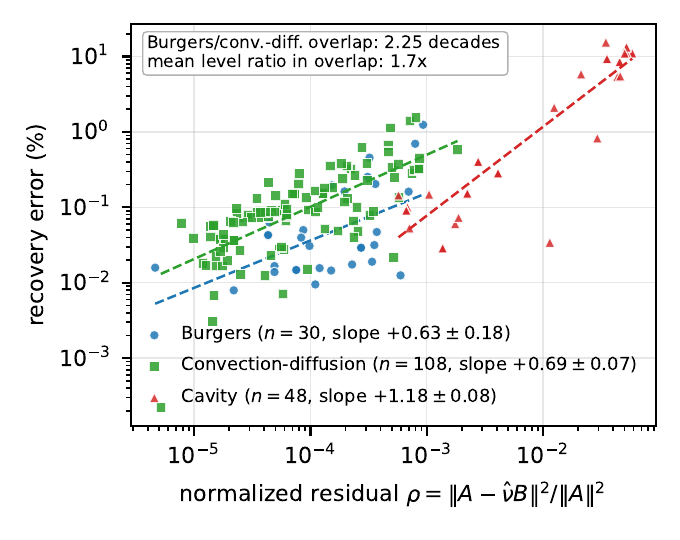}
\caption{Parameter recovery error against the normalized residual $\rho$ of Eq.~(\ref{eq:2}). Each point is one network evaluated at one training checkpoint, with $n = 30$ for Burgers, $n = 108$ for convection-diffusion and $n = 48$ for the cavity (smoothed lid, the union of the two runs of Supplementary Table~\ref{tab:S9}). Dashed lines are per-system fits drawn only over each system's own range, with slopes and standard errors in the legend. The annotation gives the overlap of the Burgers and convection-diffusion ranges of $\rho$ and the ratio of mean recovery error within it. The range of the cavity is largely disjoint from the other two, and no pooled fit is shown.}
\label{fig:2}
\end{figure}

The P\'eclet sweep from 10 to 100, with three perturbation levels, three seeds and three checkpoints, gives 108 cells from 36 networks. The pooled log-log correlation between recovery error and $\rho$ is $+ 0.72$ with slope $+ 0.69$, and it lies between $+ 0.51$ and $+ 0.77$ within each P\'eclet number and between $+ 0.59$ and $+ 0.73$ within each checkpoint. It holds both between and within networks when the dependence between checkpoints of one network is respected (Supplementary Table~\ref{tab:S8}). Across systems the exponents are near 0.65 for Burgers and convection-diffusion and near 1.20 for the cavity (Fig.~\ref{fig:2} and Supplementary Table~\ref{tab:S9}), and the cavity value replicates to within 0.013 across two runs. A fit pooled over the three systems gives a correlation of $+ 0.80$, but the cavity shares only 0.21 decades of $\rho$ with the scalar systems, so the pooled line mostly connects group means and no common relation across systems is claimed. These exponents are empirical associations for the networks and schedules studied, not derived constants. For the cavity, with six networks per run, the exponent is a within-network slope measured along training trajectories, and slopes on per-network means or final snapshots are not stable (Supplementary Table~\ref{tab:S9}).

Table~\ref{tab:4} reconstructs the angular term of Eq.~(\ref{eq:4}) on the same cells. Recomputing $A$ and $B$ from stored checkpoints reproduces Eq.~(\ref{eq:3}) with median relative discrepancies of $6.6 \times 10^{- 5}$ to $1.3 \times 10^{- 3}$. Because Eq.~(\ref{eq:3}) is an identity and the least-squares slope is linear in the fitted quantity, the measured slope equals one half plus the angular slope plus the slope of $\log K$ exactly, up to reconstruction error, and the table serves as a consistency check of the decomposition. The column ``Predicted'' omits the $\log K$ term, so its difference from the measured slope is the contribution of that term: $0.001$ for Burgers, below $0.001$ for convection-diffusion and $0.022$ for the cavity. The cavity value is larger because the cavity spans a narrow range of $\rho$ (Section~\ref{sec:5.3}, Fig.~\ref{fig:2}), over which even a variation of $\log K$ below 0.002 produces a measurable slope; it is within the standard error of the cavity exponent (0.08, Supplementary Table~\ref{tab:S9}). The analytic term contributes one half in every system, and the differences between systems are carried by the angular drift, which is $+ 0.13$ for Burgers, $+ 0.19$ for convection-diffusion and $+ 0.72$ for the cavity. The within-network slope for convection-diffusion ($+ 0.88$) is steeper than the between-network slope ($+ 0.61$) for the same reason, because $\cot\theta$ changes as a network trains.

\begin{table}[t]
\caption{The measured exponent decomposed. ``Predicted'' is $\frac{1}{2} + \dlog|\cot\theta|/\dlog\rho$, the first two terms of Eq.~(\ref{eq:4}), evaluated on the same cells as Supplementary Table~\ref{tab:S9}. The difference between ``Measured'' and ``Predicted'' is the normalization term $\dlog K/\dlog\rho$ of Eq.~(\ref{eq:4}), up to reconstruction error.}
\label{tab:4}
\centering
\small
\setlength{\tabcolsep}{4pt}
\resizebox{\columnwidth}{!}{%
\begin{tabular}{lcccc}
\toprule
System & $n$ & Measured & $\dlog|\cot\theta|/\dlog\rho$ & Predicted \\
\midrule
Burgers & 30 & $+ 0.628$ & $+ 0.129$ & $+ 0.629$ \\
Convection-diffusion & 108 & $+ 0.691$ & $+ 0.191$ & $+ 0.691$ \\
Cavity & 30 & $+ 1.201$ & $+ 0.723$ & $+ 1.223$ \\
\bottomrule
\end{tabular}}
\end{table}

Since two coefficients of one equation have different basis vectors, their exponents can also differ. Recovering $c$ and $D$ jointly in convection-diffusion costs no accuracy, with errors of $0.030 \pm 0.032\%$ and $0.076 \pm 0.088\%$. Over P\'eclet numbers from 10 to 200 the error in $c$ follows $\rho$ with slope $+ 0.85$, while the better-determined $D$, with $\norm{B_{2}}/\norm{B_{1}}$ of 14 to 19, reaches its error floor early and follows $\rho$ with slope $+ 0.25$. The ill-conditioned regime of the joint fit was not reached, because lowering $D$ sharpens the pulse and keeps the advective and diffusive basis vectors nearly orthogonal, with $\cos\left( B_{1},B_{2} \right)$ within 0.04 of zero. On Burgers the homoskedastic standard error understates the scatter across seeds by factors of 7 to 29, whereas the block bootstrap stays within a factor of 2 of it at every checkpoint.

\subsection{Self-similar blowup rate}
\label{sec:5.4}

\begin{figure}[t]
\centering
\includegraphics[width=0.82\columnwidth]{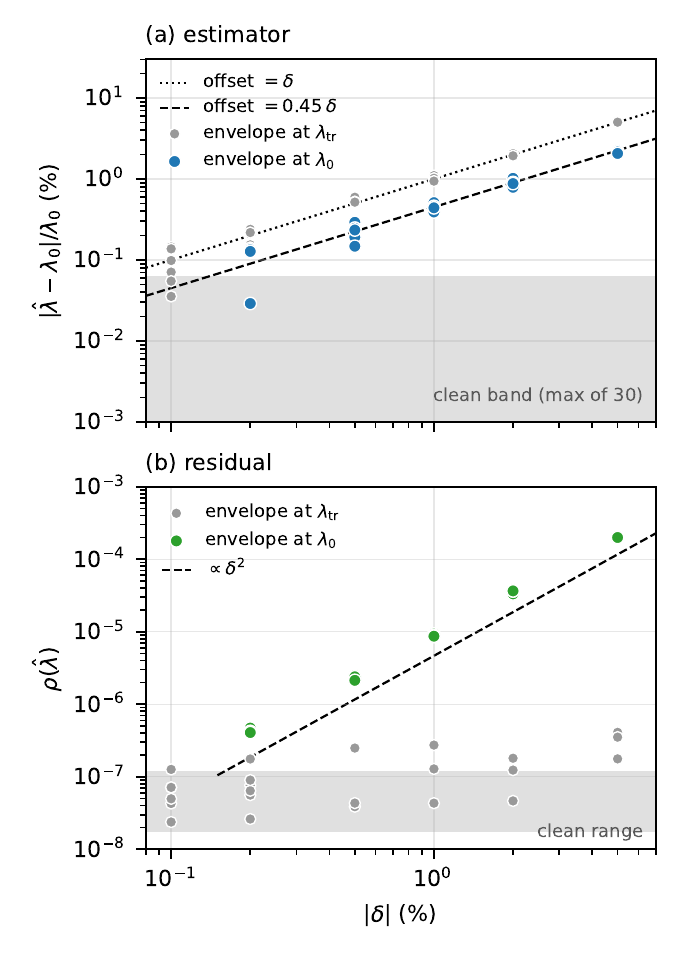}
\caption{Misspecification of the CCF blowup rate. Blue and green markers show networks with envelope and coordinate fixed at the stated rate $\lambda_{0}$ and the residual at $\lambda_{0}(1 + \delta)$ for $\delta$ of either sign (24 networks). Grey markers show networks with envelope, coordinate and residual all at $\lambda_{0}(1 + \delta)$ (24 networks). (a) Offset of the closed-form estimate from $\lambda_{0}$. The first set follows $0.45\,\delta$ (dashed) and the second follows $\delta$ (dotted); the grey band is the maximum clean offset over thirty seeds. (b) $\rho\left( \est{\lambda} \right)$ against $|\delta|$. For the first set it rises as $\delta^{2}$ (dashed) and exceeds the clean range (grey band) at every $\delta$, and for the second set it stays inside the clean range up to 2\%.}
\label{fig:3}
\end{figure}

The clean networks recover $\lambda_{0}$ with a maximum offset of 0.063\% and reach $\rho\left( \est{\lambda} \right)$ with geometric mean $5.2 \times 10^{- 8}$, three orders of magnitude below the converged benchmark networks. Two independent checks confirm the profile, since the asymptotic constant of $\omega$ is 0.938 on every seed, and the origin relation $\lambda = 2s\left( \mathcal{H}\omega \right)(0) - 2$ returns 1.179 on every seed. When only the equation is misspecified, the estimator returns a rate between those of the architecture and the equation, with $\left( \est{\lambda} - \lambda_{0} \right)/\left( \lambda_{\mathrm{tr}} - \lambda_{0} \right) = 0.45 \pm 0.06$ independent of the sign and size of $\delta$ (Fig.~\ref{fig:3}a and Supplementary Table~\ref{tab:S10}). The envelope imposes the far field of $\lambda_{0}$ and the equation that of $\lambda_{\mathrm{tr}}$, and the network converges to a profile that partly satisfies each. The offset detects every misspecification of 0.5\% or more but under-reports it by about half. The residual is more sensitive than the offset (Fig.~\ref{fig:3}b) and grows as $|\delta|^{1.93}$, symmetrically in sign, from $4.3 \times 10^{- 7}$ at 0.2\% to $2.0 \times 10^{- 4}$ at 5\%, which is 8 to 3900 times the clean geometric mean, and every misspecified network exceeds the largest clean value, so with a clean population for reference it resolves an error an order of magnitude smaller than the offset does.

When the architecture is also built from the incorrect rate, the network is consistent, $\rho\left( \est{\lambda} \right)$ stays within the clean range up to $\delta = 2\%$, and the estimator recovers $\lambda_{\mathrm{tr}}$ to 0.11\%. At $\delta = 5\%$ the profile differs from the clean one by at most 6\% in a band near $|y| \approx 1$ and by less than 1\% at the origin and beyond $|y| \approx 5$, and its residual at its own rate matches that of the clean network (Supplementary Fig.~\ref{fig:S1}). It is a well-converged approximation to a nearby equation for which no smooth solution is expected. Wang et al.~\cite{r5} resolve $\lambda_{0}$ by minimizing the attainable residual over $\lambda$ at residuals of $10^{- 8}$ to $10^{- 13}$, and at the residuals reached with Adam and L-BFGS that minimum is flat across the range tested. The estimator therefore tests the equation, the residual compared with a clean population tests agreement between equation and architecture, and a retrain-per-rate search at sufficient precision tests admissibility.

\section{Discussion and limitations}

The cavity first appears more sensitive to misspecification than Burgers' equation, but much of that difference reflects convergence, since removing the corner singularity reduces its recovery error by two orders of magnitude. Within each system, recovery accuracy is associated with $\rho$, a quantity that can be computed from the trained model. The remaining differences in exponent between systems are reproducible but not explained, and three systems are too few to separate candidate causes such as the number of coupled equations or the divergence constraint. The two-parameter results suggest that the exponent reflects how strongly each basis vector participates in the residual.

The audit complements existing tools for assessing trained PINNs and does not replace them. A posteriori certificates bound the solution error or the worst-case residual of a network \cite{r8,r9}, whereas Eq.~(\ref{eq:1}) identifies which coefficient the network is consistent with and provides no direct information about solution error. A correctly specified network can recover its stated coefficient and still carry a large solution error, as the clean cavity networks compared with the benchmark in Section~\ref{sec:5.2} illustrate, and a network with a small certified residual can still encode an incorrect coefficient, so the two assessments are complementary. Unlike inverse PINNs \cite{r1} and equation discovery \cite{r6,r7}, which estimate unknown coefficients from data, the audit needs no data and targets the discrepancy between a stated and an encoded value. The decomposition of Eq.~(\ref{eq:3}) is the classical relation between residual and parameter error in least squares \cite{r13,r14}, and its practical content lies in the measured angular term, which determines how tightly $\rho$ constrains the recovered coefficient for a given system.

The results also have implications for how PINN studies report the effect of parameter changes. At practical budgets the distance between misspecified and correctly specified Burgers networks was smaller than the distance between correctly specified networks that differ only in seed, so a single comparison against one reference network can attribute seed variation to the perturbation. Reporting a drift floor alongside such comparisons is a simple step toward the stronger baselines and more careful reporting advocated by McGreivy and Hakim \cite{r20}. Since the estimate, its bootstrap standard error and $\rho$ cost one evaluation, reporting them alongside the training loss would let others check a stated coefficient.

For self-similar formulations used in singularity discovery \cite{r5}, the results indicate that the estimator alone is not sufficient when the parameter also enters the architecture, because it then returns a compromise between the two rates. In that setting the residual compared with correctly specified networks built in the same way was the more sensitive diagnostic. When the rate was incorrect in both the equation and the architecture, neither diagnostic detected errors of up to 2\% at the residual levels reached here. Whether higher-precision training, such as the Gauss--Newton and multistage methods of \cite{r5,r17}, would make such errors detectable was not tested, and the CCF results should be read as a characterization at first-order precision.

This paper is not without its limitations. The estimator requires a residual linear in the recovered parameter, which may not hold where a rate rescales a multi-dimensional domain, as in the Boussinesq formulation of Wang et al.~\cite{r5}, although the discrete sweep remains available. The ill-conditioned multi-parameter regime was not reached, because advective and diffusive basis vectors stay nearly orthogonal in the problems studied, and the practical-budget convection-diffusion results rest on single seeds. The detection floor was calibrated on one equation, and its values should not be transferred to other systems without a new calibration. The cavity could not be trained to the residual range of the other systems, so it remains open whether the systems share one relation with different exponents.

The experiments also cover a limited set of architectures and training methods. All networks except one sine-activated variant were fully connected tanh networks, and all used equal loss weights. Adaptive loss weighting \cite{r40,r41}, Bayesian PINNs \cite{r42}, neural operators \cite{r43} and PirateNets \cite{r44} were not tested. The estimator and $\rho$ are computed from the interior residual alone and do not depend on loss weights directly, but weighting changes the trained network, so its effect on recovery accuracy remains to be measured. The two training budgets used here are single points in a wide range of possible schedules, and the relation between $\rho$ and recovery error may differ for other optimizers or much longer training.

\section{Conclusions}

A PINN trained with a misspecified coefficient solves the equation it was given, so its training loss does not reliably indicate the error, and at practical budgets the raw loss can even favor the misspecified model through a scaling effect that the normalized residual does not share. The coefficient a frozen network encodes is nevertheless recoverable in closed form from one evaluation, with an accuracy that depends on the normalized residual; the empirical exponents, 0.63 to 1.20 across three equations, decompose into one half plus an angular drift. On Burgers' equation a single-model rule based on the spatial block-bootstrap standard error flagged 3 of 300 correctly specified networks (an observed false-alarm rate of 1\%), flagged every misspecification of 2\% or more at the practical budget, and after convergence flagged two thirds of 1\% misspecifications with no false alarm among 30 clean networks, which does not establish a zero false-alarm probability. When the parameter also enters the architecture, as for the CCF blowup rate, the estimator under-reports a misspecified equation by about half while $\rho$ grows approximately quadratically and resolves a 0.2\% error, and a rate that is incorrect in both equation and architecture was not detected at the precision reached with first-order training. A low training loss therefore does not show that a PINN encodes its documented coefficient, whereas the closed-form estimate, its bootstrap standard error and the normalized residual together indicate what the model encodes and how reliable that estimate is.

\clearpage
\onecolumn

\setcounter{table}{0}
\setcounter{figure}{0}
\renewcommand{\thetable}{S\arabic{table}}
\renewcommand{\thefigure}{S\arabic{figure}}
\captionsetup[table]{name=Supplementary Table}
\captionsetup[figure]{name=Supplementary Fig.}
\captionsetup[table]{skip=8pt}
\setlength{\textfloatsep}{22pt plus 4pt minus 2pt}
\setlength{\intextsep}{22pt plus 4pt minus 2pt}
\setlength{\floatsep}{18pt plus 4pt minus 2pt}
\renewcommand{\arraystretch}{1.15}

\begin{center}
{\large\bfseries Supplementary material for ``Parameter misspecification in physics-informed neural networks: A closed-form audit and its limits''}
\end{center}
\vspace{1em}

\section*{Experimental configurations}

Table~\ref{tab:S1} lists the configuration of every experiment. All networks are fully connected with tanh activation, except one sine-activated variant in the architecture comparison, and are trained with Adam at learning rate $10^{- 3}$, followed where stated by L-BFGS. Experiments on Burgers' equation, the lid-driven cavity and convection-diffusion use single precision, Glorot uniform initialization and the L-BFGS-B implementation with $\mathit{ftol} = 2.2 \times 10^{- 15}$, $\mathit{gtol} = 10^{- 5}$ and at most 15,000 iterations. The detection-floor and CCF experiments use double precision, Glorot normal initialization and the PyTorch L-BFGS implementation with $\mathit{ftol} = 10^{- 15}$ and $\mathit{gtol} = 10^{- 10}$, for 30,000 and 20,000 iterations respectively. Burgers' equation is solved on $x \in [- 1,1]$, $t \in [0,1]$ with $u(x,0) = - \sin\pi x$ and $u( \pm 1,t) = 0$. For the cavity the pressure is not gauge-fixed, the two momentum residuals are stacked into one vector of 20,000 entries, and evaluation points within 0.02 of the walls are excluded. The one-dimensional convection-diffusion problem uses inflow Dirichlet data and a Gaussian initial pulse. The experiments ran on NVIDIA T4 and RTX 4000 Ada GPUs, about 380 GPU-hours in total.

\begin{table*}[!htbp]
\caption{Configuration of each experiment. Practical budget refers to the checkpoint at the end of the Adam phase and converged to the checkpoint after L-BFGS. Collocation counts are interior, boundary and initial points. $^{a}$The architecture of the single-seed two-dimensional convection-diffusion run was not recorded.}
\label{tab:S1}
\centering
\footnotesize
\setlength{\tabcolsep}{4pt}
\begin{tabular}{>{\raggedright\arraybackslash}p{2.3cm}>{\raggedright\arraybackslash}p{2.6cm}>{\raggedright\arraybackslash}p{3.9cm}>{\raggedright\arraybackslash}p{1.4cm}>{\raggedright\arraybackslash}p{2.3cm}>{\raggedright\arraybackslash}p{3.3cm}}
\toprule
Experiment & Problem & Network, collocation & Precision & Checkpoints & Networks \\
\midrule
Loss and solution difference & Burgers; cavity $\Re = 100,400$; 2D conv.-diff. & Burgers $[20] \times 3$; cavity $[50] \times 4$; 2D conv.-diff. not recorded$^{a}$ & single & practical & 3 seeds per level (conv.-diff. 1) \\
Loss ordering & cavity $\Re = 400$, standard lid, $\delta = 50\%$ & $[50] \times 4$ & single & 25k, 50k, 100k Adam; L-BFGS & 3 per arm \\
External benchmark & cavity $\Re = 400$ vs. 600 & $[64] \times 4$, both lids & single & 60k and 300k & 3 per arm, 4 groups \\
Drift and exponents, Burgers & Burgers & $[20] \times 3$; 2540, 80, 160 & single & Adam stages; L-BFGS & 7 clean, 7 at $\delta = 50\%$ (drift); 5 clean, 5 at $\delta = 50\%$ (exponent fit) \\
Residual quality, cavity & cavity, smoothed and standard lid & $[64] \times 4$; 4000, 600 & single & Adam stages; L-BFGS & 6 per run, 2 runs \\
P\'eclet sweep & 1D conv.-diff., $\Pe = 10$ to 200 & $[20] \times 3$ & single & 3 stages & 36; 15 two-parameter runs \\
Detection floor & Burgers & $[20] \times 3$; 2540, 80, 160 & double & 20k Adam; 30k L-BFGS & 300 clean (30 converged); 6 per level \\
Blowup rate & CCF profile & $[128] \times 4$; 4097-point quadrature grid & double & 60k Adam; 20k L-BFGS & 30 clean; 24 + 24 misspecified \\
\bottomrule
\end{tabular}
\end{table*}

\section*{Loss and solution difference at the practical budget}

Table~\ref{tab:S2} gives the training loss and solution difference for every perturbation level at the practical budget. The column ``exact'' is the difference between the true solutions of the equation at $\lambda(1 + \delta)$ and at $\lambda$. For Burgers the true solutions differ by 0.25, 0.49, 0.96, 2.3, 4.2 and 7.3\% across the six levels, against PINN differences of 0.6, 1.6, 1.3, 2.8, 4.8 and 8.0\%. The convection-diffusion training losses of $10^{- 5}$ to $10^{- 6}$ approach the single-precision floor, and those rows are single-seed.

\begin{table}[!htbp]
\caption{Training loss and solution difference at the practical budget (mean $\pm$ standard deviation over three seeds for the cavity and Burgers; convection-diffusion single-seed except at $\delta = 200\%$). Training losses vary strongly across seeds, in some cases with standard deviation exceeding the mean, while $L^{2}$ differences are more stable. The column ``exact'' is the relative $L^{2}$ difference between the true solutions at $\lambda(1 + \delta)$ and at $\lambda$, from the Cole--Hopf solution for Burgers and from the analytic Gaussian pulse ($D = 0.01$, $|c| = 0.335$, $\sigma_{0} = 0.1$) for convection-diffusion, an idealization of the experimental configuration. It is not available for the cavity, where reference solutions exist only at $\Re = 100$ and 400.}
\label{tab:S2}
\centering
\footnotesize
\setlength{\tabcolsep}{3.5pt}
\begin{tabular}{llccc}
\toprule
Problem & $\delta$ & Training loss & $L^{2}$ difference & Exact \\
\midrule
Burgers ($\nu$) & 0 & $0.011 \pm 0.010$ & -- & -- \\
 & 5\% & $0.0006 \pm 0.0006$ & $0.006 \pm 0.001$ & 0.003 \\
 & 10\% & $0.006 \pm 0.008$ & $0.016 \pm 0.002$ & 0.005 \\
 & 20\% & $0.0009 \pm 0.0008$ & $0.013 \pm 0.002$ & 0.010 \\
 & 50\% & $0.003 \pm 0.003$ & $0.028 \pm 0.002$ & 0.023 \\
 & 100\% & $0.0007 \pm 0.0001$ & $0.048 \pm 0.001$ & 0.042 \\
 & 200\% & $0.0001 \pm 0.0001$ & $0.080 \pm 0.000$ & 0.072 \\
\addlinespace
Cavity, $\Re = 100$ & 0 & $0.0033 \pm 0.0006$ & -- & -- \\
 & 5\% & $0.0127 \pm 0.0066$ & $0.241 \pm 0.090$ & -- \\
 & 10\% & $0.0051 \pm 0.0013$ & $0.131 \pm 0.070$ & -- \\
 & 20\% & $0.0119 \pm 0.0055$ & $0.248 \pm 0.139$ & -- \\
 & 50\% & $0.0089 \pm 0.0035$ & $0.279 \pm 0.014$ & -- \\
 & 100\% & $0.0080 \pm 0.0025$ & $0.415 \pm 0.112$ & -- \\
\addlinespace
Cavity, $\Re = 400$ & 0 & $0.0088 \pm 0.0003$ & -- & -- \\
 & 5\% & $0.0089 \pm 0.0008$ & $0.320 \pm 0.189$ & -- \\
 & 10\% & $0.0082 \pm 0.0002$ & $0.256 \pm 0.179$ & -- \\
 & 20\% & $0.0084 \pm 0.0008$ & $0.401 \pm 0.151$ & -- \\
 & 50\% & $0.0089 \pm 0.0016$ & $0.563 \pm 0.117$ & -- \\
 & 100\% & $0.0071 \pm 0.0005$ & $0.706 \pm 0.123$ & -- \\
\addlinespace
Convection-diffusion ($D$) & 0 & $6 \times 10^{- 6}$ & -- & -- \\
 & 5\% & $8 \times 10^{- 6}$ & 0.019 & 0.015 \\
 & 10\% & $8 \times 10^{- 6}$ & 0.035 & 0.030 \\
 & 20\% & $1.0 \times 10^{- 5}$ & 0.063 & 0.058 \\
 & 50\% & $1.1 \times 10^{- 5}$ & 0.148 & 0.130 \\
 & 100\% & $2.1 \times 10^{- 5}$ & 0.260 & 0.224 \\
 & 200\% & $(2.5 \pm 1.5) \times 10^{- 5}$ & $0.405 \pm 0.002$ & 0.355 \\
\bottomrule
\end{tabular}
\end{table}

\section*{Seed drift}

Table~\ref{tab:S3} compares the drift floor with the perturbation effect for Burgers at $\delta = 50\%$. The ratio of the two rises monotonically from 0.79 after 20k Adam steps to 1.49 after L-BFGS. Over the same interval the drift falls by a factor of 3.8 and the perturbation distance by a factor of 2, while the standard deviation of the perturbation distance decreases tenfold, from 0.0334 to 0.0032. Even after L-BFGS the effect is 0.85 standard deviations of the drift, and the largest drift distance (0.0376) exceeds the smallest perturbation distance (0.0227). For the cavity the separation is complete, with a mean drift of 0.182 against a mean perturbation distance of 1.331, and a largest drift of 0.281 against a smallest perturbation distance of 0.907.

\begin{table}[!htbp]
\caption{Drift floor and perturbation effect for Burgers at $\delta = 50\%$, seven clean and seven misspecified networks. Drift is the relative $L^{2}$ distance between clean pairs (21 pairs, mean $\pm$ standard deviation), and the effect is the distance between each misspecified and each clean network (49 pairs). The effect size $d = \left( \overline{\text{effect}} - \overline{\text{drift}} \right)/\mathrm{sd}\left( \text{drift} \right)$ is the separation in units of seed-to-seed spread.}
\label{tab:S3}
\centering
\small
\begin{tabular}{lcccc}
\toprule
Checkpoint & Drift & Effect & Ratio & $d$ \\
\midrule
20k Adam & $0.067 \pm 0.046$ & $0.054 \pm 0.033$ & 0.79 & $- 0.30$ \\
50k Adam & $0.053 \pm 0.040$ & $0.042 \pm 0.027$ & 0.79 & $- 0.28$ \\
100k Adam & $0.030 \pm 0.018$ & $0.031 \pm 0.011$ & 1.03 & $+ 0.05$ \\
L-BFGS & $0.018 \pm 0.010$ & $0.027 \pm 0.003$ & 1.49 & $+ 0.85$ \\
\bottomrule
\end{tabular}
\end{table}

\section*{Loss-based checks}

Table~\ref{tab:S4} lists six checks based on the loss or on quantities derived from it, evaluated at the practical budget. Evaluating the residual against the stated equation gives a $3.3 \times$ elevation at $\Re = 100$ but requires a reference baseline that is not available in practice. Three misspecified models disagree with one another (standard deviation 0.470) more than they differ from the clean baseline (0.175), and an ensemble trained at $\pm 2$, $\pm 5$ and $\pm 10\%$ around the stated value does not isolate the misspecified model. Fitting the parameter by gradient descent on the frozen network returns a value that follows its initialization, whereas the closed form, which is the exact minimizer of the same objective, has no initialization. The divergence of the velocity is higher for the misspecified model at $\Re = 100$ and 150 and lower at 400 and 800. Loss elevation against a clean baseline is $2.7 \times$ at $\Re = 100$ and inverted at 400, and 38\% of clean networks at $\Re = 100$ exceed the threshold $\tau = 0.01$.

\begin{table*}[!htbp]
\caption{Loss-based checks at the practical training budget, cavity unless stated, with the number of trained networks in parentheses. The entry marked with an asterisk was recorded only as a summary statistic.}
\label{tab:S4}
\centering
\small
\begin{tabular}{>{\raggedright\arraybackslash}p{5.6cm}>{\raggedright\arraybackslash}p{6.0cm}>{\raggedright\arraybackslash}p{4.6cm}}
\toprule
Check (networks) & Statistic & Value \\
\midrule
Residual against stated equation (1 + 1) & misspecified/clean residual, $\Re = 100$ & $3.3 \times$ \\
Same-parameter ensemble (3) & member s.d. / difference from clean & 0.470 / 0.175 \\
Parameter-jitter ensemble (3 per $\Re$) & misspecified/ensemble deviation, $\Re = 100,150,800$ & 0.81, 0.44, 0.57 \\
Gradient recovery on frozen network (5 initializations)$^{*}$ & recovered $\Re$ & tracks initialization \\
Incompressibility (1 per $\Re$) & $\nabla \cdot \mathbf{u}$, misspecified vs. clean & higher at 100, 150; lower at 400, 800 \\
Loss against clean baseline (3 seeds; 21 clean) & elevation at $\Re = 100$; at $\Re = 400$; clean above threshold & $2.7 \times$; inverted; 38\% \\
\bottomrule
\end{tabular}
\end{table*}

Table~\ref{tab:S5} shows that misspecification of the cavity passes unnoticed by the loss in both directions. Six of eight perturbations around $\Re = 100$ leave the loss at or near the baseline, and at $\Re = 90$ a 10\% reduction gives a 62.5\% solution difference at a loss of 0.00324 against 0.00334. For Burgers no perturbation in either direction goes unnoticed. Table~\ref{tab:S6} shows that the solution difference is large for all five architectures tested, from 8.7k to 133k parameters, while the loss elevation ranges from $1.03 \times$ for the $[64] \times 3$ network, which also has the largest solution difference, to $2.0 \times$ for the $[256] \times 3$ network.

\begin{table}[!htbp]
\caption{Misspecification of the cavity in both directions (true $\Re = 100$), single seed per level. The two perturbations visible in the loss, $\Re = 105$ and 200, carry the highest training losses in the sweep.}
\label{tab:S5}
\centering
\small
\begin{tabular}{lccc}
\toprule
$\Re$ used & Training loss & $L^{2}$ difference & Invisible to loss \\
\midrule
50 & 0.00392 & 0.134 & yes \\
75 & 0.00389 & 0.274 & yes \\
90 & 0.00324 & 0.625 & yes \\
100 & 0.00334 & 0 & reference \\
105 & 0.01066 & 0.085 & no \\
110 & 0.00861 & 0.090 & yes \\
120 & 0.00691 & 0.250 & yes \\
150 & 0.00376 & 0.431 & yes \\
200 & 0.01113 & 0.560 & no \\
\bottomrule
\end{tabular}
\end{table}

\begin{table}[!htbp]
\caption{Architecture dependence of loss detectability for the cavity at $\delta = 50\%$, mean $\pm$ standard deviation over three seeds.}
\label{tab:S6}
\centering
\footnotesize
\setlength{\tabcolsep}{3.5pt}
\begin{tabular}{lcccc}
\toprule
Architecture & Parameters & Clean loss & Misspecified loss & $L^{2}$ difference \\
\midrule
$[128] \times 5$ tanh & 66.8k & $0.0042 \pm 0.0010$ & $0.0066 \pm 0.0013$ & $0.405 \pm 0.057$ \\
$[64] \times 3$ tanh & 8.7k & $0.0147 \pm 0.0005$ & $0.0151 \pm 0.0004$ & $1.302 \pm 0.410$ \\
$[256] \times 3$ tanh & 133k & $0.0065 \pm 0.0007$ & $0.0129 \pm 0.0004$ & $0.265 \pm 0.017$ \\
$[64] \times 8$ tanh & 29.5k & $0.0064 \pm 0.0031$ & $0.0076 \pm 0.0026$ & $0.486 \pm 0.265$ \\
$[128] \times 5$ sin & 66.8k & $0.0047 \pm 0.0010$ & $0.0075 \pm 0.0002$ & $0.266 \pm 0.163$ \\
\bottomrule
\end{tabular}
\end{table}

\section*{External benchmark}

Clean cavity models trained at $\Re = 400$ and misspecified models trained at $\Re = 600$ but documented as 400 were compared with the $\Re = 400$ centerline benchmark used in the main text (Table~\ref{tab:S7}). The misspecified models are worse by factors of 1.2 to 1.8 in $u$ and 1.0 to 1.7 in $v$, but the clean models already carry errors of 10 to 26\% in $u$ and 20 to 63\% in $v$. The standard-lid group carries about $2.6 \times$ the clean $u$ error and $3.2 \times$ the clean $v$ error of the smoothed-lid group trained under otherwise identical settings. Over the full domain the clean and misspecified predictions differ by 28\%, so the seventeen-point centerline comparison registers only a small part of the discrepancy.

\begin{table*}[!htbp]
\caption{Relative $L^{2}$ error against the $\Re = 400$ centerline tables of Ghia et al.\ cited in the main text, three networks per arm. Misspecified models are trained at $\Re = 600$ and documented as $\Re = 400$. The benchmark uses a uniform lid, so the standard-lid group is the like-for-like comparison, and part of the deviation of the smoothed-lid groups is attributable to the boundary condition.}
\label{tab:S7}
\centering
\small
\resizebox{\textwidth}{!}{%
\begin{tabular}{lcccccc}
\toprule
Group & Clean, 400, $u$ & Clean, 400, $v$ & Misspecified, 600, $u$ & Misspecified, 600, $v$ & Ratio, $u$ & Ratio, $v$ \\
\midrule
Smoothed lid, 60k & 0.099 & 0.197 & 0.180 & 0.325 & 1.81 & 1.65 \\
Standard lid, 60k & 0.265 & 0.632 & 0.462 & 0.933 & 1.74 & 1.47 \\
Smoothed lid, 300k & 0.112 & 0.232 & 0.137 & 0.239 & 1.22 & 1.03 \\
Smoothed lid, additional seeds & 0.127 & 0.268 & 0.193 & 0.342 & 1.52 & 1.28 \\
\bottomrule
\end{tabular}}
\end{table*}

\section*{Sensitivity to the evaluation domain}

For the cavity, recovery is stable for wall-strip widths $h_{w} \in [0.01,0.05]$. With $h_{w} = 0$ the recovered Reynolds numbers span 413 to 738 against true values of 400 and 600, because the lid corners contaminate the fit, and with $h_{w} = 0.10$ they span 365 to 593 because genuine flow structure is excluded. The default $h_{w} = 0.02$ lies inside the stable range.

\section*{Recovery error and normalized residual}

Table~\ref{tab:S8} repeats the regression of recovery error on $\rho$ in the P\'eclet sweep under treatments that respect the dependence between the three checkpoints of each network, including a bootstrap that resamples networks. The relation holds between and within networks, with an effective sample size of 36. Table~\ref{tab:S9} gives the per-system fits. For Burgers and convection-diffusion the slopes on per-network means agree with the pooled slopes. For the cavity they do not ($+ 0.22$ in run 1 and $- 0.01$ on final snapshots), because with six networks per run the cavity exponent is measured along training trajectories. A fit pooled over the three systems gives a correlation of $+ 0.797$ over 3.87 decades of $\rho$, but the cavity shares only 0.21 decades with the other systems, and the between-group correlation ($+ 0.978$ on three points) dominates the within-group value ($+ 0.684$). Where Burgers and convection-diffusion overlap, over 2.25 decades, convection-diffusion has a recovery error 1.7 to 2.6 times larger at equal $\rho$, depending on how snapshots are weighted. A standard-lid cavity fit with exponent $+ 3.02$ over 0.85 decades joins two clusters and is excluded.

\begin{table}[!htbp]
\caption{Recovery error against $\rho$ in the P\'eclet sweep under treatments that respect the dependence between snapshots of one network. The bootstrap resamples networks, not cells.}
\label{tab:S8}
\centering
\small
\begin{tabular}{lccc}
\toprule
Treatment & $n$ & Correlation & Slope \\
\midrule
All cells, pooled & 108 & $+ 0.715$ & $+ 0.691$ \\
Per-network means & 36 & $+ 0.778$ & $+ 0.606$ \\
Final snapshot only & 36 & $+ 0.677$ & $+ 0.645$ \\
Adam snapshots only & 72 & $+ 0.709$ & $+ 0.708$ \\
Within-network (demeaned) & 108 & $+ 0.682$ & $+ 0.882$ \\
Bootstrap over networks, 95\% interval & 36 & $[0.63,0.80]$ & $[0.53,0.89]$ \\
\bottomrule
\end{tabular}
\end{table}

\begin{table*}[!htbp]
\caption{Per-system power-law fits of recovery error against $\rho$ over all snapshots, with the ordinary least-squares standard error of the slope and a 95\% interval from a bootstrap that resamples networks. ``Per-net.'' is the slope on per-network means and ``final'' uses only the last snapshot of each network. The cavity uses the smoothed lid; its two runs differ in training budget and are also pooled.}
\label{tab:S9}
\centering
\small
\resizebox{\textwidth}{!}{%
\begin{tabular}{lcccccccc}
\toprule
System & $n$ & Networks & Slope & s.e. & Bootstrap 95\% & $r$ & Per-net. & Final \\
\midrule
Burgers & 30 & 10 & $+ 0.63$ & 0.18 & $[+ 0.31, + 1.12]$ & $+ 0.55$ & $+ 0.69$ & $+ 0.28$ \\
Convection-diffusion & 108 & 36 & $+ 0.69$ & 0.07 & $[+ 0.53, + 0.89]$ & $+ 0.71$ & $+ 0.61$ & $+ 0.65$ \\
Cavity, run 1 & 18 & 6 & $+ 1.19$ & 0.17 & $[+ 1.04, + 1.53]$ & $+ 0.87$ & $+ 0.22$ & $- 0.39$ \\
Cavity, run 2 & 30 & 6 & $+ 1.20$ & 0.08 & $[+ 1.08, + 1.54]$ & $+ 0.94$ & $+ 0.79$ & $+ 0.77$ \\
Cavity, pooled & 48 & 12 & $+ 1.18$ & 0.08 & $[+ 1.07, + 1.39]$ & $+ 0.92$ & $+ 0.92$ & $- 0.01$ \\
\bottomrule
\end{tabular}}
\end{table*}

\section*{Two-parameter recovery}

For convection-diffusion with both coefficients unknown the residual $u_{t} + cu_{x} - Du_{xx}$ gives $A = u_{t}$, $B_{1} = - u_{x}$ and $B_{2} = u_{xx}$. Over fifteen runs spanning perturbations of both coefficients, $c$ is recovered to $0.030 \pm 0.032\%$ and $D$ to $0.076 \pm 0.088\%$, the same as for $D$ recovered alone with $c$ known, and joint and single estimates differ by at most 0.004 percentage points. Over P\'eclet numbers from 10 to 200, which span 3.66 decades of $\rho$, the error in $c$ follows $\rho$ with correlation $+ 0.902$ and slope $+ 0.851$, while the error in $D$ follows it with correlation $+ 0.424$ and slope $+ 0.247$ and its mean stays between 0.04 and 0.23\% across the P\'eclet levels (per-run values from 0.0005 to 0.44\%). The ratio $\norm{B_{2}}/\norm{B_{1}}$ ranges from 14 to 19. Across the P\'eclet range $\operatorname{cond}\left( M^{\top}M \right)$ varies only from $1.95 \times 10^{2}$ to $3.47 \times 10^{2}$ and $\cos\left( B_{1},B_{2} \right)$ stays within 0.04 of zero, becoming more orthogonal at high P\'eclet number. Mean $\log_{10}\rho$ falls from $- 3.47$ at $\Pe = 10$ to $- 4.60$ at $\Pe = 100$.

\section*{Calibration of the bootstrap}

On Burgers the homoskedastic standard error of the estimator understates the observed scatter across seeds by factors of 28.5, 11.4, 12.0 and 7.1 at successive checkpoints. The spatial block bootstrap gives ratios of observed to estimated scatter between 0.55 and 1.75 throughout. On the converged double-precision Burgers networks the clean signed offsets have a standard deviation of 0.202\% against a median bootstrap standard error of 0.199\%. In the two-parameter runs the smallest errors in $c$, about $10^{- 6}$, are a few single-precision units in the last place.

\section*{CCF setup and additional results}

The network represents $\omega = \Theta'$, which is odd and decays as $|y|^{- 1/(1 + \lambda)}$, in the coordinate $q = \left( 1 + y^{2} \right)^{- 1/\left( 2(1 + \lambda) \right)}$ with a fixed odd prefactor and decay envelope. $\Theta$ and $\mathcal{H}\Theta$ are integrated from $\omega$ and $\mathcal{H}\omega$ starting at the origin, so every Hilbert transform is of a decaying function, and it is computed by principal-value quadrature on a sinh-compactified grid of 4097 points with an analytic tail whose constant is the network's own value at $q = 0$. The loss is the mean squared residual scaled by $\left( 1 + y^{2} \right)^{- a/2}$ with $a = \lambda/(1 + \lambda)$ and weighted by $1/\left( 1 + (y/20)^{2} \right)$, plus $10\,\left( \Theta''(0) - 1 \right)^{2}$ to fix the scaling symmetry. Training uses 60k Adam steps with cosine decay and 20k L-BFGS iterations on a $[128] \times 4$ network in double precision. The rate enters the residual, the far-field exponent of the envelope and the coordinate. Because the residual is linear in $\lambda$, its squared norm is quadratic in $\lambda$ for a frozen profile and fixed evaluation weights, so $\rho\left( \est{\lambda} \right)$ is computed exactly as the vertex of the parabola through three sweep points.

The main experiment uses $\delta \in \{ - 2, - 1, - 0.5,0.2,0.5,1,2,5\}\%$ with three seeds each. The thirty clean networks recover $\lambda_{0}$ with mean signed offset $- 0.015\%$ and standard deviation 0.038\%, and their $\rho\left( \est{\lambda} \right)$ ranges from $1.7 \times 10^{- 8}$ to $1.2 \times 10^{- 7}$. Table~\ref{tab:S10} gives the results per level. The fraction of the perturbation recovered by the estimator is $0.47 \pm 0.03$ for negative and $0.44 \pm 0.08$ for positive $\delta$, with standard error 0.013 over the 21 networks with $|\delta| \geq 0.5\%$. The residual takes the values $2.3 \times 10^{- 6}$ at $\pm 0.5\%$, $8.9 \times 10^{- 6}$ at $\pm 1\%$ and $3.6 \times 10^{- 5}$ at $\pm 2\%$.

\begin{figure*}[!htbp]
\centering
\includegraphics[width=\textwidth]{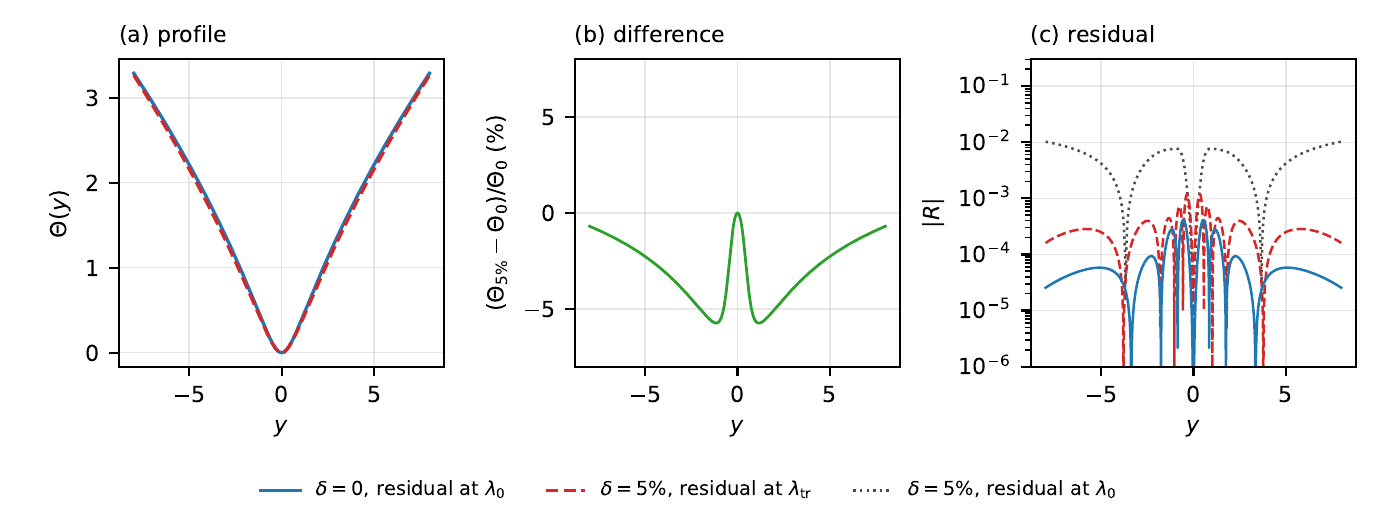}
\caption{CCF profiles for a correctly specified network and for a network with envelope, coordinate and residual all built from $\lambda_{0}(1 + \delta)$, $\delta = 5\%$, seed 0. (a) Profiles $\Theta(y)$ on $|y| < 8$. (b) Relative difference between the two profiles, which is confined to a band near $|y| \approx 1$ and is below 1\% at the origin and beyond $|y| \approx 5$. (c) Magnitude of the profile-equation residual for the clean network at $\lambda_{0}$, for the misspecified network at its training rate, and for the misspecified network at $\lambda_{0}$. The first two are indistinguishable in magnitude and the third is thirty times larger.}
\label{fig:S1}
\end{figure*}

In the second set, with envelope, coordinate and residual all built from $\lambda_{0}(1 + \delta)$ for $\delta$ from 0.1 to 5\%, the geometric means of $\rho\left( \est{\lambda} \right)$ are 5.7, 7.0, 7.5, 11.5, 10.1 and $29 \times 10^{- 8}$, inside the clean range up to 2\% and five times above it at 5\% (Spearman correlation $+ 0.43$, $p = 0.001$). Fig.~\ref{fig:S1} compares the profile at $\delta = 5\%$ with the clean profile.

\begin{table*}[!htbp]
\caption{CCF blowup rate with envelope and coordinate built from the stated rate $\lambda_{0}$ and the residual at $\lambda_{0}(1 + \delta)$, three seeds per level. Thirty clean networks calibrate the bands (maximum offset 0.063\%, maximum $\rho\left( \est{\lambda} \right) = 1.21 \times 10^{- 7}$, geometric mean $5.2 \times 10^{- 8}$). ``Min.\ offset'' is the smallest offset in percent, and ``fraction'' is $\left( \est{\lambda} - \lambda_{0} \right)/\left( \lambda_{\mathrm{tr}} - \lambda_{0} \right)$, the share of the perturbation recovered by the estimator. The column $\rho/\rho_{\mathrm{clean}}$ is the geometric mean over seeds relative to the clean geometric mean. The last two columns count seeds whose offset, and whose $\rho\left( \est{\lambda} \right)$, exceed the clean maximum.}
\label{tab:S10}
\centering
\small
\begin{tabular}{lcccccc}
\toprule
$\delta$ & $n$ & Min.\ offset & Fraction & $\rho/\rho_{\mathrm{clean}}$ & Offset $>$ max & $\rho >$ max \\
\midrule
$- 2\%$ & 3 & 0.877 & $0.47 \pm 0.04$ & 713 & 3/3 & 3/3 \\
$- 1\%$ & 3 & 0.440 & $0.45 \pm 0.01$ & 173 & 3/3 & 3/3 \\
$- 0.5\%$ & 3 & 0.234 & $0.48 \pm 0.01$ & 45 & 3/3 & 3/3 \\
$+ 0.2\%$ & 3 & 0.029 & $0.48 \pm 0.29$ & 8 & 2/3 & 3/3 \\
$+ 0.5\%$ & 3 & 0.148 & $0.42 \pm 0.15$ & 44 & 3/3 & 3/3 \\
$+ 1\%$ & 3 & 0.391 & $0.44 \pm 0.06$ & 169 & 3/3 & 3/3 \\
$+ 2\%$ & 3 & 0.789 & $0.45 \pm 0.05$ & 672 & 3/3 & 3/3 \\
$+ 5\%$ & 3 & 2.065 & $0.43 \pm 0.02$ & 3948 & 3/3 & 3/3 \\
\bottomrule
\end{tabular}
\end{table*}

\end{document}